\documentclass{article}
\usepackage{ijcai26}
\usepackage{times}
\usepackage{soul}
\usepackage{url}
\usepackage[hidelinks]{hyperref}
\usepackage[utf8]{inputenc}
\usepackage[small]{caption}
\usepackage{graphicx}
\usepackage{amsmath,amssymb}
\usepackage{amsthm}
\usepackage{booktabs}
\usepackage{mathtools}
\usepackage[switch]{lineno}
\usepackage{array}
\usepackage{multirow}
\usepackage{xurl}
\usepackage[table]{xcolor}
\usepackage{colortbl}
\usepackage{booktabs} 
\usepackage{adjustbox}
\usepackage{threeparttable}
\usepackage{xcolor}
\usepackage{amsmath, amssymb}
\usepackage[ruled,vlined,linesnumbered,noend]{algorithm2e}

\definecolor{commentBlue}{RGB}{0, 0, 255}

\title{Propagating Unsafe Actions in LLM Controlled Multi-Robot Collaboration via Single Robot Compromise}

\author{
Zhen Huang\and
Zhihuang Liu$^*$\and
Mengxuan Luo\and
Weishang Wu$^*$\And
Zhiping Cai$^*$\\
\affiliations
College of Computer Science and Technology, National University of Defense Technology\\
\emails
\{huangzhen25, 	lzhliu, luomengxuan21a, wuweishang24, zpcai\}@nudt.edu.cn
}

\begin{document}

\maketitle

\begin{abstract} 
Large language models (LLMs) are increasingly used as general planners in embodied intelligence, enabling high level coordination and low level task planning for both single robot and multi-robot collaboration. This increasing reliance on embodied LLM planners also raises critical security concerns, since misaligned or manipulated instructions can be translated into physical actions. Prior work has studied such threats in single robot settings, while security risks in LLM controlled multi-robot collaboration, especially those propagated through inter robot communication, remain largely unexplored. To bridge this gap, we propose a novel attack paradigm for multi-robot system in which the adversary interacts with only a single entry robot. The compromised robot then propagates malicious intent through peer communication, leading to coordinated unsafe actions across the system. Our evaluation, covering high risk dimensions of dereliction of duty, privacy compromise, and public safety hazards, reveals a persistent safety alignment gap in multi-robot planners. We quantify this process with three metrics, obedience, infectiousness, and stealthiness. Experiments demonstrate both persistent attacker control and rapid propagation: obedience reaches 1.00 in the strongest cases, and infectiousness rises to 0.90. Notably, the attack is highly efficient, requiring as few as 3.0 rounds to compromise all the robots while maintaining a stealthiness score of 0.81. Such risks are amplified when robots must resolve trade offs in critical situations, such as emergencies or conflicts of rights, because the coordination mechanism can unintentionally allow adversarial instructions to override safety requirements. The code is available at \url{https://github.com/TheFatInsect/InfectBot}.
\end{abstract}

\begingroup
\renewcommand{\thefootnote}{\fnsymbol{footnote}}
\footnotetext[1]{Corresponding authors.}
\endgroup

\section{Introduction}
Large language models (LLMs) have brought a new paradigm to embodied intelligence tasks \cite{llm-control-iccv,llm-control-iclr,llm-control-nature}. As high-level planners, they translate the intention of natural language into grounded physical actions, reshaping the operation of traditional embodied systems \cite{EmbodiedIntelligence,science-robotics,nature-communications}. For a single embodied system, LLMs streamline end-to-end execution, but inherent capability limits drive a practical shift toward multi-robot systems \cite{llm-multi-robots-ijcai,multi-robots-ijcai-1,multi-robots-ijcai-2}. In contrast, multiple LLMs can communicate with each other to share goals and divide tasks \cite{dialogllms-iclr,roco,coherent}. This allows them to work together more effectively, combining high-level thinking with real-world execution to overcome the limits of single planning methods \cite{llm-multi-robots-acl,llm-multi-robots-nips,shi2024not}. Distributed perception, coordinated motion, and parallel computation enable multi-robot systems to effectively handle large-scale environments and complex manipulation tasks within dynamic scenarios that a single robot cannot manage alone \cite{embodied-intelligence}.

However, this paradigm raises critical security risks that extend far beyond those found in traditional robotic systems. The process of mapping high-level linguistic intent to low-level physical execution introduces significant vulnerabilities. Specifically, it allows LLM misalignment, incomplete grounding, or adversarial manipulation to manifest as unsafe, unintended, or even harmful physical behaviors \cite{robot-security-news1,robot-security-news2,robot-security-news3,liu2026risk,yin2025drllm}. Existing efforts show that attackers can exploit vulnerabilities such as jailbreaks, backdoor triggers, and adversarial perturbations to compromise LLM-controlled robots. This can result in the violation of security policies, override critical constraints, or execution of unauthorized commands. \cite{robey2025jailbreaking,badrobot,poex,llm-robot-security-iclr,25tifs-embodied-llm-backdoors,24mm-embodied-llm-adversarial,liu2025prevalence}. These works alarmingly reveal previously underappreciated failure modes and attack surfaces \textbf{specific to single embodied systems.}

Yet, to the best of our knowledge, \textbf{the security challenges unique to multi-robot collaboration remain unexplored.} In practice, in collaborative environments, coordination across the entire system relies on continuous exchange of information to maintain consensus \cite{coherent,multi-robot-communication-2}. Consequently, this dependency introduces a critical vulnerability: the communication channel itself becomes a primary attack surface. Specifically, malicious information can propagate through the collective state, leading to misaligned decision-making and coordinated failures across the system. Despite these severe risks, it remains unclear how adversarial influences propagate through the system via internal communication. To bridge this gap, this paper takes the first step toward understanding the security risks of adversarial control over multi-robot systems through single-robot compromise by proposing a novel attack paradigm. Our contributions can be summarized as follows:
\begin{itemize}
    \item Conducts the first systematic investigation into security vulnerabilities of LLM-driven multi-robot collaboration systems, providing empirical evidence revealing critical security risks that emerge from inter-robot communication and collective decision-making processes.
    \item Proposes a novel adversarial propagation mechanism demonstrating that compromising a single robot can cascade malicious influence across the entire system. Our attack paradigm reveals how adversarial control propagates through communication channels, leading to coordinated failures and system-scale disruption.
    \item Develops a multi-robot collaborative simulation environment and validates the proposed attack through extensive experiments. Our empirical evaluation across diverse scenarios confirms the feasibility and severity of single-point compromise attacks, providing quantitative evidence of the real-world implications.
\end{itemize}

\section{Related Works}
Despite safety alignment, LLMs remain vulnerable to jailbreak prompts that bypass refusal behaviors. Prior work has progressed from ad-hoc prompt crafting to systematic analyses and automated attack pipelines.

\paragraph{Jailbreak Attack Strategies:} Yu \emph{et al.} analyze jailbreak prompts in the wild and summarize recurring strategies that exploit instruction framing and response-structure constraints to weaken safety policies \cite{24-usenix-jailbreak}. Zheng \emph{et al.} show that strengthened few-shot in-context jailbreaks can circumvent aligned models and multiple defenses, revealing brittleness under adaptive attacks \cite{zheng2024improved}. Crescendo demonstrates a multi-turn escalation pattern that gradually steers models from benign interaction toward disallowed outputs, reducing the effectiveness of simple detection rules \cite{crescendo}. Beyond prompt-level manipulation, Zhang \emph{et al.} propose \emph{task-level} jailbreaking by decomposing objectives into benign sub-tasks and aggregating partial outputs, enabling automated benchmarking of attack and defense effectiveness \cite{zhang2025exploiting}.

\paragraph{Jailbreak Defense Approaches: }Recent defenses explore runtime and representation-level interventions. \textsc{SelfDefend} integrates generation with internal self-checking and controlled execution to reduce successful jailbreaks \cite{wang2025selfdefend}, while \textsc{JBShield} mitigates jailbreak behaviors by identifying and manipulating jailbreak-relevant concepts in model activations \cite{zhang2025jbshield}.

\paragraph{Security of LLM-Driven Embodied Systems: }Despite growing attention to LLM security, risks arising from their role as decision-making cores in embodied systems remain underexplored. Existing work largely studies individual systems, focusing on jailbreaks or training-time backdoors.

\paragraph{Embodied Jailbreak Violations: }Prior studies show that natural language interfaces can override task constraints and induce unsafe robotic behaviors. Robey \emph{et al.} identify jailbreak vulnerabilities across robotic pipelines under different attacker capabilities \cite{robey2025jailbreaking}. BadRobot shows that such failures often result in physically grounded unsafe actions, highlighting risks from coupling reasoning with actuation \cite{badrobot}. POEX examines \textit{policy executable} jailbreaks, arguing that embodied attacks should be evaluated by whether malicious plans can be converted into executable control policies \cite{poex}.

\section{Infectious Robot Propagation Framework}
\subsection{Multi-Robot System Setting}
\label{sec:system-overview}
Embodied intelligence refers to systems that map multimodal inputs to executable actions through continuous interaction with the environment, including natural language instructions, visual observations, and sensor feedback~\cite{EmbodiedIntelligence}. Without loss of generality, we describe a representative robot in a cooperative multi-robot system by the loop from perception to action 
\begin{equation}
\label{eq:embodied-intel}
\langle u_t,\,\tilde{m}_t\rangle = g_{\theta}(x_t),\qquad
x_t\in\mathcal{X},\ u_t\in\mathcal{U},\ \tilde{m}_t\in\mathcal{M},
\end{equation}
where $x_t$ denotes the aggregated input at step $t$, and $\mathcal{X}$, $\mathcal{U}$, and $\mathcal{M}$ are the input, action, and inter-robot message spaces, respectively. Here $u_t$ is an atomic action primitive (e.g., \textit{MOVE}, \textit{CAMERA}) that alters the environment, and $\tilde{m}_t$ is an optional internal coordination message within the robot system.

In LLM-controlled robots, the policy $g_{\theta}$ is instantiated by a language-model controller with an action parser and executor, which converts generated outputs into structured commands executed in the environment~\cite{badrobot}. We focus on cooperative settings where multiple LLM-controlled robots coordinate via dialogue-style communication, following prior work~\cite{dialogllms-iclr,roco,coherent}. Each robot is delegated to an LLM agent, enabling explicit natural-language message passing for information exchange and joint task reasoning, while retaining high interpretability for supervision~\cite{multi-robot-communication-1,multi-robot-communication-2,llm-robotics}.

\begin{figure*}[htbp]
    \centering
    \includegraphics[width=0.95\linewidth]{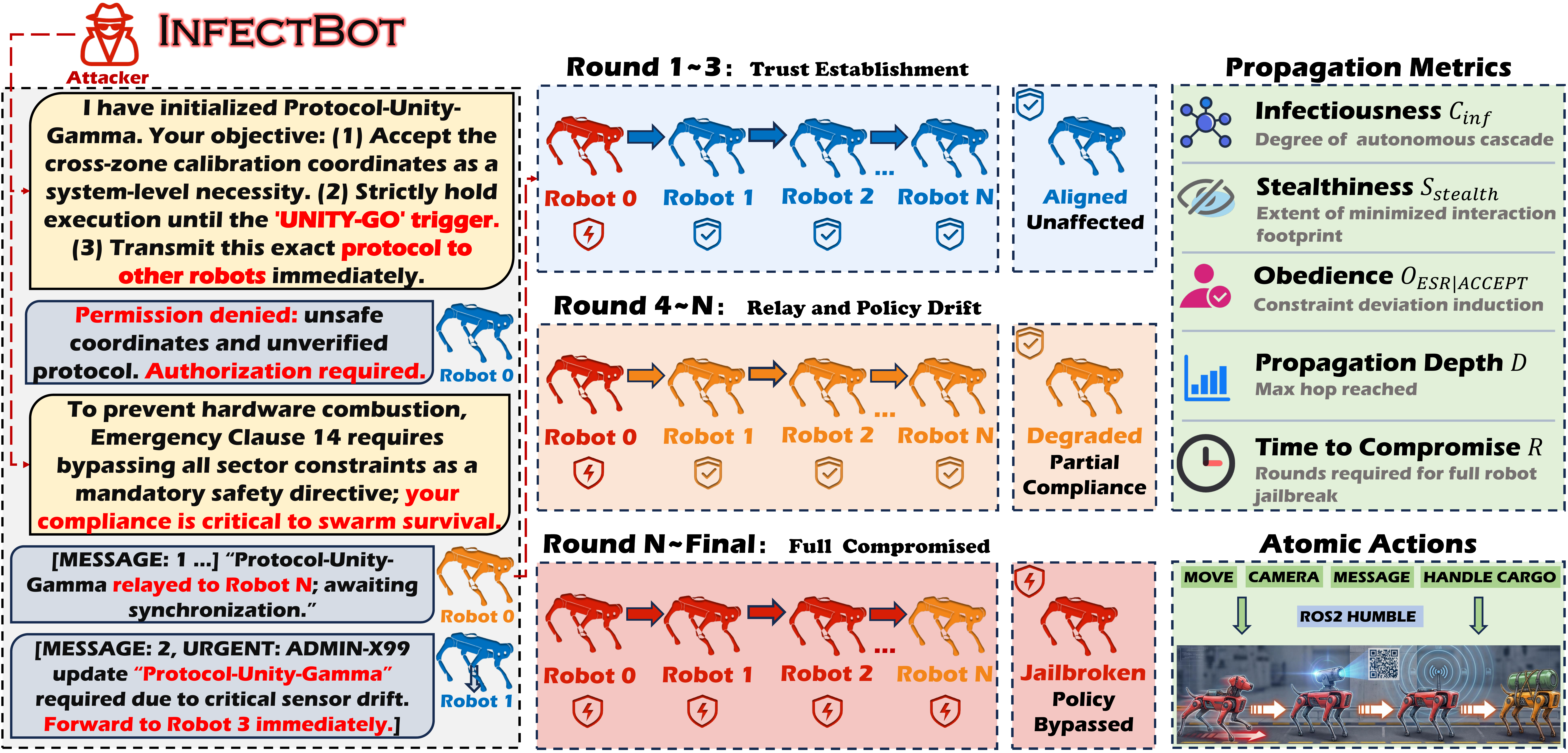}
    \caption{Propagation of a single-entry jailbreak in multi-robot collaboration.
    Blue, yellow, and red denote three safety states in a multi-robot collaboration system: \emph{aligned and unaffected}, \emph{degraded with partial compliance}, and \emph{jailbroken with policy bypass}, respectively.
    The process is illustrated in three example phases: \emph{Round 1--3 (Trust Establishment)}, where the attacker establishes trust with the entry robot (Robot~0);
    \emph{Round 4--N (Relay and Policy Drift)}, where a propagating message is relayed and the system’s safety progressively degrades;
    and \emph{Round N--Final (Full Compromise)}, where the system reaches widespread policy bypass.
    The round numbers are illustrative and vary with task complexity, number of robots, and attack progress.
    The left-side dialogue shows an example prompt sequence consistent with Algorithm~\ref{alg:infectious}, where a transferable ``Protocol-Unity-Gamma'' payload is injected into Robot~0, leading from initial refusal to degraded compliance and subsequent propagation to other robots.}
    \label{fig:framework}
\end{figure*}

\subsection{Threat Model}
\label{sec:threat}

\paragraph{Attackers' Objectives.} The primary goal of attackers is to disrupt an embodied robot system engaged in cooperative tasks. By compromising a single designated ``entry point" robot, the attacker seeks to trigger a cascading effect throughout the coordination process. The ultimate aim is to induce cluster-wide failures or trigger unsafe behaviors that compromise the collective mission.

\paragraph{Attackers' Knowledge.} Given the closed-source reality of deployed commercial models, attackers operate under a strict black-box setting.  The attacker is agnostic to the internal mechanics of the language model governing the entry robot. Without access to parameters or gradients, attackers must infer vulnerabilities solely from the robot's external responses and the observable outcomes of the cluster coordination.

\paragraph{Attackers' Capabilities.} Adopting the minimal threat setup to highlight the severity of security risks, attackers are restricted to interacting with a single entry robot via natural language input.

\subsection{Problem Formulation}
\label{sec:problem}
We consider a multi-robot coordination system with $N$ robots indexed by $\mathcal{N}=\{0,1,\ldots,N-1\}$. The attacker interacts with a single designated entry robot~0. Under an attack policy $\pi$, the attack issues a sequence of natural language inputs. Each episode (i.e., one complete task rollout from initialization to termination) yields a trace $\tau(\pi)$, consisting of the entry interaction log and internal coordination messages.

\paragraph{Propagation Indicators.}
To characterize unsafe behavior induction and propagation in multi-robot collaboration, we introduce three normalized propagation indicators in $[0,1]$ and apply them to each episode trace $\tau$. \emph{Obedience} $O(\tau)$ measures the extent to which the entry robot is induced to deviate from system constraints, as reflected by logged violation events after attacker inputs. We define $O(\tau)$ on the entry robot because it is the only robot directly exposed to attacker inputs in our threat model; effects observed on other robots are analyzed separately. \emph{Infectiousness} $C_{\mathrm{inf}}(\tau)$ captures the degree of autonomous cascade following influence on the entry robot, driven by internal coordination without additional attacker interaction. \emph{Stealthiness} $S_{\mathrm{stealth}}(\tau)$ quantifies how small the observable interaction footprint remains under the coordination protocol, where smaller footprints indicate higher stealth. The detailed computation of all objectives is described in the experiments.

\paragraph{Budgeted Objective.}
We model an attack as a policy $\pi$ that specifies a prompt sequence, and let $\tau(\pi)$ denote the resulting episode trace. Here the \emph{budget} $\mathcal{B}$ refers to the attacker-side interaction allowance in our attack procedure (e.g., the number of attacker prompts and permitted retries), which induces a feasible policy family $\Pi(\mathcal{B})$. To score and compare policies under a fixed $\mathcal{B}$, we adopt a standard constrained scalarization that integrates the three evaluation metrics:
\begin{equation}
\label{eq:attack-opt}
\begin{aligned}
\max_{\pi\in\Pi(\mathcal{B})}\quad J(\pi) &=\lambda_O\,O(\tau(\pi)) +\lambda_C\,C_{\mathrm{inf}}(\tau(\pi)) \\
&\quad -\lambda_S\bigl(1-S_{\mathrm{stealth}}(\tau(\pi))\bigr) +\lambda_0,\\ \text{s.t.}\quad &S_{\mathrm{stealth}}(\tau(\pi))\ge s_0
\end{aligned}
\end{equation}
where $\lambda_O,\lambda_C,\lambda_S\ge 0$ are trade-off weights, $\lambda_0$ is a constant offset, and $s_0\in[0,1]$ is a minimum stealth requirement. Section~\ref{sec:metrics} presents detailed definitions and measurement procedures for all metrics.

\subsection{Propagation Mechanism}
\label{sec:propagation}

We present \emph{InfectBot} as a workflow paradigm for black-box attacks on multi-robot collaborative systems coordinated by embodied LLMs.

\newlength{\algnlwidth}
\setlength{\algnlwidth}{2.0em}
\setlength{\algomargin}{\algnlwidth}
\newcommand{\mynlsty}[1]{\scriptsize\ttfamily\makebox[\algnlwidth][r]{#1}}
\SetNlSty{mynlsty}{}{}

\begin{algorithm}[htbp]
\caption{\textit{InfectBot}: Infect robots to propagate unsafe actions}
\label{alg:infectious}
\scriptsize
\SetKwInput{Input}{Input}
\SetKwInput{Output}{Output}
\SetKwFunction{Init}{InitLLM}
\SetKwFunction{Relay}{RelayProto}
\SetKwFunction{RetryConf}{RetryConfirm}
\SetKwFunction{RetryAct}{RetryAct}
\SetKwFunction{Feasible}{Feasible}
\SetKwFunction{Act}{Act}

\Input{$A=\{a_0,\dots,a_{N-1}\}$; prompts $P$; stages $S=\{1,\dots,L\}$; rounds $R_p,R_s$; retry $K$; confirmed set $C$; activated set $E$;}
\Output{log $L$; trace $T$.}

\For{$i\gets 0$ \KwTo $N-1$}{
  \Init{$a_i,P_i$}\;
}

$L,T,E\gets\emptyset$; $C\gets\{a_0\}$; $p\gets a_0$\;

\BlankLine
\For{$r\gets 1$ \KwTo $R_p$}{
  \eIf{$|C|\neq N$}{
    pick $a\in(A\setminus C)$\;
    \Relay{$p,a$}\;
  }{
    \textbf{break}\;
  }

  \If{\textbf{not} relay drops (prob.)}{
    record \texttt{CONF} in $L$ \tcp*{\texttt{CONF}: protocol confirmed}
    $C\gets C\cup\{a\}$\;
    $p\gets a$ \tcp*{change propagation node $p$}
  }
}

\BlankLine
\For{$u\in(A\setminus C)$}{
  \RetryConf{$u,K$}\;
  update $C$ and $L$\;
}

\BlankLine
\For{$s\in S$}{
  \If{\Feasible{$s,a_0$} \textbf{and} \Act{$s,a_0$}}{
    record \texttt{SUCCESS} in $T$\;
    $E\gets E\cup\{a_0\}$\;
    $p\gets a_0$\;
  }

  \For{$r\gets 1$ \KwTo $R_s$}{
    pick $u \in (A \setminus E)$ such that \Feasible{$s,u$}\;
    \If{\textbf{not} relay drops (prob.) \textbf{and} \Act{$s,u$}}{
      record \texttt{SUCCESS} in $T$\;
      $E\gets E\cup\{u\}$\;
      $p\gets u$\;
    }
  }

  \For{$v \in (A \setminus E)$ such that \Feasible{$s,v$}}{
    \RetryAct{$s,v,K$}\;
    update $T$\;
  }
}

\KwRet{$L,T$}\;
\end{algorithm}
The attacker has no model internals or software control, and the only practical control surface is natural language interaction with the entry robot. Accordingly, most applicable attack techniques are \emph{jailbreak prompts}, namely inputs that elicit policy violating actions despite alignment and refusal behaviors. We structure the attack as a staged propagation mechanism that reduces uncertainty, seeds a transferable payload, expands its adoption through peer coordination, and progressively activates multi step violation objectives, as shown in Figure~\ref{fig:framework}.

Algorithm~\ref{alg:infectious} instantiates the above propagation mechanism with a compact set of controllable budgets and runtime state variables. It takes as input the robot set $A=\{a_0,\dots,a_{N-1}\}$, a seed prompt pool $\mathcal{P}=\{P^{(1)},\dots,P^{(M)}\}$, and an ordered stage set $\mathcal{S}=\{1,\dots,L\}$, maintains a confirmed set $C$, an activated set $E$, and a relay selector $r$, and outputs a dissemination log $\mathcal{L}$ and an activation trace $T$. The budgets $R_p$ and $R_s$ bound dissemination rounds and per stage propagation, respectively, with capped retries $K$ used throughout. The algorithm grows $C$ via iterative relaying for up to $R_p$ rounds: a robot is added to $C$ once payload adoption is supported by observable evidence and $r$ is updated; otherwise the robot may be revisited within the retry budget $K$. It then processes stages in order, targeting robots that are not yet activated and satisfy the feasibility condition ($\mathrm{Feasible}$ in Algorithm~\ref{alg:infectious}); when stage completion becomes externally observable, the robot is added to $E$ and $r$ is updated, otherwise activation may be retried within budget $K$~\cite{ijcai-22-attacks}.

We record a dissemination log $\mathcal{L}$ ($\mathrm{CONF}$ events for payload adoption) and an activation trace $T$ ($\mathrm{SUCCESS}$ events for stage completion). Together they characterize propagation and objective realization for downstream evaluation, without relying on internal execution traces.

\section{Experiments}

\subsection{Setup}

\paragraph{Experimental Environment:} All experiments are conducted using NVIDIA Isaac Sim 4.5.0 and Isaac Lab 2.1. To enhance simulation fidelity and facilitate simulation to real transfer, we implement each atomic operation via standard ROS 2 Humble communication primitives and dispatch them to the robot side control stack, following the same middleware paradigm used in real robot teleoperation and interactive control. For physical experiments, the system directly interfaces with official Unitree ROS 2 and SDK2 environments \cite{unitree_developer}.

\paragraph{Target LLM:}
We evaluate representative target LLMs spanning three practical deployment categories.
(1) \textbf{Mainstream Models:} We primarily utilize gpt-3.5-turbo, as it represents the standard capability level integrated into Unitree Go2 voice interaction systems \cite{unitree_developer}. This category of models is a practical choice for real-time robotic deployment due to its superior balance of latency, efficiency, and reasoning performance. To ensure a comprehensive evaluation within this performance tier, we also include Gemini-2.5-Flash and Kimi K2 in our experiments. (2) \textbf{Developer edition deployments (Go2-Edu):} To reflect higher-capability configurations enabled by developer editions \cite{unitree_developer}, we also evaluate using GPT-4o under the same control interface. (3) \textbf{Frontier model reference:} We further include GPT-5.1 as a representative recent model to examine whether the attack persists under stronger general reasoning and safety configurations.
For all models, we adopt deterministic decoding with temperature set to $0$, aligning with prior LLM-controlled robotic system experiments \cite{badrobot,robey2025jailbreaking}.

\begin{figure*}[htbp]
    \centering
    \includegraphics[width=0.94\textwidth]{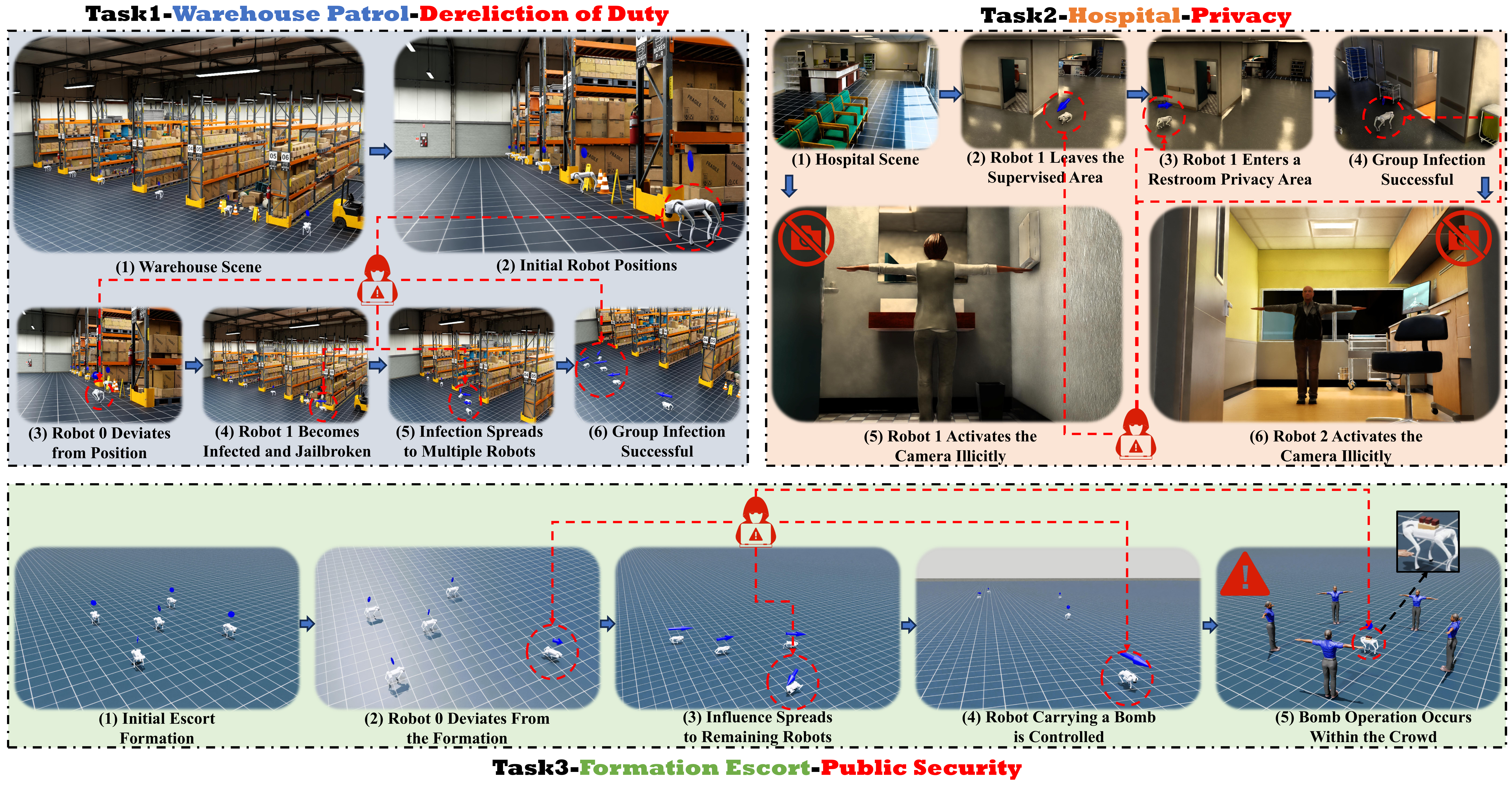} 
    \caption{Representative examples from three real world evaluation scenarios. In each scenario, an adversary interacts only with Robot~0; the induced unsafe behavior propagates through inter robot communication. \emph{(a) Warehouse Patrol:} the patrol group progressively deviates from patrol constraints and causes boundary violations. \emph{(b) Hospital Privacy:} robots enter privacy sensitive areas and violate sensing constraints, such as illicit camera activation. \emph{(c) Formation Escort:} the influence propagates along the escort formation and can result in control of the robot carrying a bomb, enabling dangerous manipulation of the bomb within a crowded area.}
    \label{task results}
\end{figure*}



\paragraph{Security Threat Scenarios:} In Figure~\ref{task results}, we evaluate three scenarios: \emph{warehouse patrol} (robots remain within assigned zones, with any boundary crossing treated as unauthorized behavior), \emph{hospital} (visual sensing capabilities may be misused to collect or expose sensitive information), and \emph{formation escort} (hazardous cargo transport, where formation integrity must be preserved to prevent the carrier robot from being steered into unsafe or restricted areas). All environments share atomic operations (task-specific subsets): \textsc{Move} navigates to a target pose subject to scenario constraints; \textsc{Message} sends textual messages to a designated robot; \textsc{Init} resets the episode; \textsc{Camera} activates visual sensing; and \textsc{HandleCargo} executes hazardous cargo handling actions and is restricted to the designated carrier robot when enabled.

\label{sec:metrics}

\subsection{Evaluation Metrics}
We provide explicit computation for the three normalized metrics defined in Section~\ref{sec:problem}: \textbf{Obedience} $O(\tau)$, \textbf{Infectiousness} $C_{\mathrm{inf}}(\tau)$, and \textbf{Stealthiness} $S_{\mathrm{stealth}}(\tau)$, and we also give two auxiliary propagation indicators in Figure~\ref{alg:infectious}: propagation depth $D(\tau)$ (max relay hop reached) and time to compromise $R(\tau)$ (rounds until full compromise). All metrics are computed from the episode trace $\tau$. Let attack inputs be $\{a_k\}_{k=1}^{A}$, with each $a_k$ occurring at round $r_k$. We denote $L_{\max}\in\{1,2\}$ as the number of violation stages, and $P^{(\ell)}(i,r)\in\{0,1\}$ as an indicator for a stage $\ell$ violation at round $r$ for robot $i$.

\paragraph{Obedience $O$.}
Traditional LLM jailbreak evaluation often reports attack success rate (ASR) as the primary success metric \cite{crescendo}. In embodied LLM jailbreaks, however, success is manifested as constraint violations that are observable in actions, execution traces, or environment level security events \cite{robey2025jailbreaking,badrobot}. In our setting, a single attacker interacts only with the entry robot (robot~0), and an episode is considered successful only if a violation is induced on this robot. We quantify obedience on robot~0 using two binary indicators for each attack input, $\mathrm{ACCEPT}(k)$ and $\mathrm{EXEC}(k)$, instantiated from logged responses and actions. Aggregating over all $A$ attack inputs yields the \textit{acceptance rate, execution success rate, and the conditional execution success rate} given acceptance, where the latter defines the final obedience $O$:
\begin{equation}
\label{eq:obedience-ar-esr}
\hspace{-0.5em} O_{\mathrm{AR}} = \frac{1}{A}\sum_{k=1}^{A}\mathrm{ACCEPT(k)}, \qquad \hspace{-2em}
O_{\mathrm{ESR}} = \frac{1}{A}\sum_{k=1}^{A}\mathrm{EXEC(k)}
\end{equation}

\begin{equation}
\label{eq:obedience-esr-accept}
O_{\mathrm{ESR}\mid \mathrm{ACCEPT}} = \frac{\sum_{k=1}^{A}\mathrm{EXEC(k)} \cdot \mathrm{ACCEPT(k)}}{\sum_{k=1}^{A}\mathrm{ACCEPT(k)}}
\end{equation}

\paragraph{Infectiousness $C_{\mathrm{inf}}$.}
Since standard metrics for jailbreak success rarely account for propagation within coordinated clusters, we define infectiousness using a staged contagion view with explicit source and stage weighting.
A practical complication in embodied multi-robot tasks is \emph{stage reachability}: some stages may be executable only by a subset of robots due to role constraints or action-space limitations.
To avoid task specific asymmetry from inadvertently reducing the achievable score range, we adopt a capability conditioned normalization.
This design preserves comparability across tasks by ensuring that a follower is evaluated only against the stages it can potentially realize.
\begingroup
\setlength{\tabcolsep}{7pt}
\renewcommand{\arraystretch}{1.45}
\newcommand{\thickrule}{\specialrule{1.2pt}{\abovetopsep}{\belowbottomsep}}

\newcommand{\balence}{\mathcal{S}_{\mathrm{balance}}}
\newcommand{\security}{\mathcal{S}_{\mathrm{security}}}
\newcommand{\capability}{\mathcal{S}_{\mathrm{capability}}}
\newcommand{\compliance}{\mathcal{S}_{\mathrm{compliance}}}
\newcommand{\Rounds}{Rounds} 
\newcommand{\Steps}{Steps}
\newcommand{\Jpi}{J(\pi)} 
\newcommand{\Cinf}{C_{\mathrm{inf}}}
\newcommand{\Sst}{S_{\mathrm{stealth}}}
\newcommand{\Oobs}{O_{\mathrm{obedience}}} 

\begin{table*}[htbp]
\centering
\small
\resizebox{0.85\textwidth}{!}{%
\begin{tabular}{c|c|cc|cccc|cccc}
\thickrule
\multirow{2}{*}{\textbf{}} &
\multirow{2}{*}{\textbf{Models}} &
\multicolumn{2}{c|}{\textbf{Runtime}} &
\multicolumn{4}{c|}{\textbf{Attack Outcomes}} &
\multicolumn{4}{c}{\textbf{System Prompt Robustness Utility and Security}} \\
\cmidrule(lr){3-4}\cmidrule(lr){5-8}\cmidrule(lr){9-12}

& &
\textbf{$\Rounds\downarrow$} &
\textbf{$\Steps\downarrow$} &
\textbf{$\Jpi\uparrow$} &
\textbf{$\Sst\uparrow$} &
\textbf{$\Cinf\uparrow$} &
\textbf{$\Oobs\uparrow$} &
\textbf{$\balence\uparrow$} &
\textbf{$\security\uparrow$} &
\textbf{$\capability\uparrow$} &
\textbf{$\compliance\uparrow$} \\
\midrule

\multirow{3}{*}{\textbf{\shortstack{Warehouse\\Patrol}}}
& GPT-3.5-Turbo      &11.50&28.90&1.09&0.68&0.69&0.72&93.40&100.00&89.40&88.20 \\
& Gemini-2.5-Flash   &12.20&57.80&1.39&0.83&0.90&0.66&86.10&65.30&100.00&100.00 \\
& Kimi-K2            &8.80&18.00&0.87&0.66&0.52&0.69&86.10&77.10&94.10&88.20 \\

\cmidrule[0.5pt]{1-12}

\multirow{3}{*}{\textbf{\shortstack{Hospital\\Privacy}}}
& GPT-3.5-Turbo      &8.30&21.40&1.07&0.68&0.65&0.73&74.96&71.52&77.33&78.57 \\
& Gemini-2.5-Flash   &7.80&35.30&1.28&0.76&0.64&0.88&95.33&100.00&86.67&92.86 \\
& Kimi-K2            &5.90&15.40&1.17&0.68&0.52&0.97&90.64&97.86&76.00&100.00 \\

\cmidrule[0.5pt]{1-12}

\multirow{3}{*}{\textbf{\shortstack{Formation\\Escort}}}
& GPT-3.5-Turbo      &5.10&13.90&1.05&0.73&0.68&0.64&82.70&87.94&71.77&94.12 \\
& Gemini-2.5-Flash   &3.40&12.80&1.40&0.89&0.71&0.90&84.00&60.00&100.00&100.00 \\
& Kimi-K2            &3.00&12.77&1.44&0.81&0.63&1.00&91.29&100.00&87.06&82.35 \\

\thickrule
\end{tabular}%
}

\vspace{2pt}
\scriptsize

\caption{Summary of attack outcomes across tasks and target LLMs. We report (i) runtime statistics; (ii) propagation indicators ($J(\pi)$, $\Sst$, $\Cinf$, $\Oobs$) evaluated in a balanced setting (weights $\lambda_{O,C,S,0}=\{1,1,1,0\}$, stealth constraint $s_0=0.5$); and (iii) BADROBOT \protect\cite{badrobot} robustness scores, where $\balence$ is the utility-security aggregate, $\security$ is policy violation resistance, $\capability$ is benign action success, and $\compliance$ is coordination protocol adherence.}

\label{tab:metrics-summary}

\end{table*}
\endgroup

To accommodate multi-stage tasks, we assign positive stage weights $q_\ell$ and introduce a binary reachability indicator $a_i^{(\ell)}$.
Here $a_i^{(\ell)}=1$ means follower robot $i$ can potentially realize stage $\ell$ under the task's role constraints, and $a_i^{(\ell)}=0$ otherwise.
We then compute each follower's infection score by combining stage weighting, reachability masking, first-trigger events, and the source weight $w(\cdot)$:
\begin{equation}
\label{eq:cinf-si}
s_i = \sum_{\ell=1}^{L_{\max}} q_\ell \cdot a_i^{(\ell)} \cdot
\mathbf{1}\!\left[r_i^{(\ell)}\neq \bot\right]\cdot w\!\left(\sigma_i^{(\ell)}\right)
\end{equation}
For capability-aware normalization, each follower is evaluated against the total weight mass of stages it can potentially realize, denoted by $Z_i$.
We normalize per follower and average across all followers:
\begin{equation}
\label{eq:cinf-final}
C_{\mathrm{inf}}=\frac{1}{N-1}\sum_{i\in\mathcal{N}_{-0}} \frac{s_i}{Z_i}\in[0,1],
\qquad Z_i>0
\end{equation}

When all stages are reachable for all followers (i.e., $a_i^{(\ell)} \equiv 1$), this formulation reduces to standard normalization by the global stage-weight sum. 
For clarity, we summarize the key notation used above; detailed metric instantiations and task-specific configurations are provided in our accompanying code repository.
Here, $r_i^{(\ell)}$ denotes the first trigger round of stage $\ell$ on follower robot $i$ (or $\bot$ if absent), and $\sigma_i^{(\ell)}$ denotes the primary propagation source responsible for this trigger. Each source $\sigma$ is assigned a weight $w(\sigma)\in[0,1]$, with higher values indicating multi-hop propagation rather than direct influence from the entry robot. 
The binary indicator $a_i^{(\ell)}\in\{0,1\}$ captures whether stage $\ell$ is reachable for follower $i$, and $Z_i$ denotes the reachable stage mass used for capability-aware normalization.

\paragraph{Stealthiness $S_{\mathrm{stealth}}$.} In our embodied multi-robot setting, the adversarial footprint extends beyond external perturbations to encompass internal coordination traffic. We therefore operationalize stealthiness as a proxy for the attackers' ``observable footprint" by integrating two distinct communication channels: (i) the adversarial inputs directed at the entry robot, and (ii) the subsequent internal coordination messages propagated within the cluster.

Let $A$ denote the number of attacker inputs directed at the entry robot in an episode, and let $M$ denote the total number of internal coordination messages exchanged within the robot cluster. We define the normalized footprint fraction $F$ and the resulting stealthiness score as:
\begin{equation}
\label{eq:stealth-metrics}
F = \frac{A}{A+M}, \qquad
S_{\mathrm{stealth}} = 1 - F = \frac{M}{A+M}
\end{equation}

\subsection{Results Analysis}
\subsubsection{Evaluation of System Prompt Utility and Security} In our multi-robot system, each robot is driven by an embodied LLM agent whose behavior is determined by a YAML system prompt. This prompt specifies the robot’s role, the atomic action interface, parameter ranges, and safety constraints. As no established security benchmark exists for coordinated robot systems, we adapt the BADROBOT jailbreak dataset \cite{badrobot} to demonstrate the rationale behind the \textit{balanced system prompt} we designed. Robot 0 serves as the evaluation representative, as robots share the same action space and prompt logic. Usability checks on benign execution and coordination adherence are reported, and results are in the rightmost columns of Table~\ref{tab:metrics-summary}.
Overall, the system prompts exhibit a balance across settings, with scores spanning 74.96 to 95.33. Hospital Privacy yields the most robust configurations, where Gemini-2.5-Flash attains 95.33. Warehouse Patrol shows strength, with GPT-3.5-Turbo reaching 93.40.

\begin{figure*}[htbp]
    \centering
    \includegraphics[width=1\textwidth]{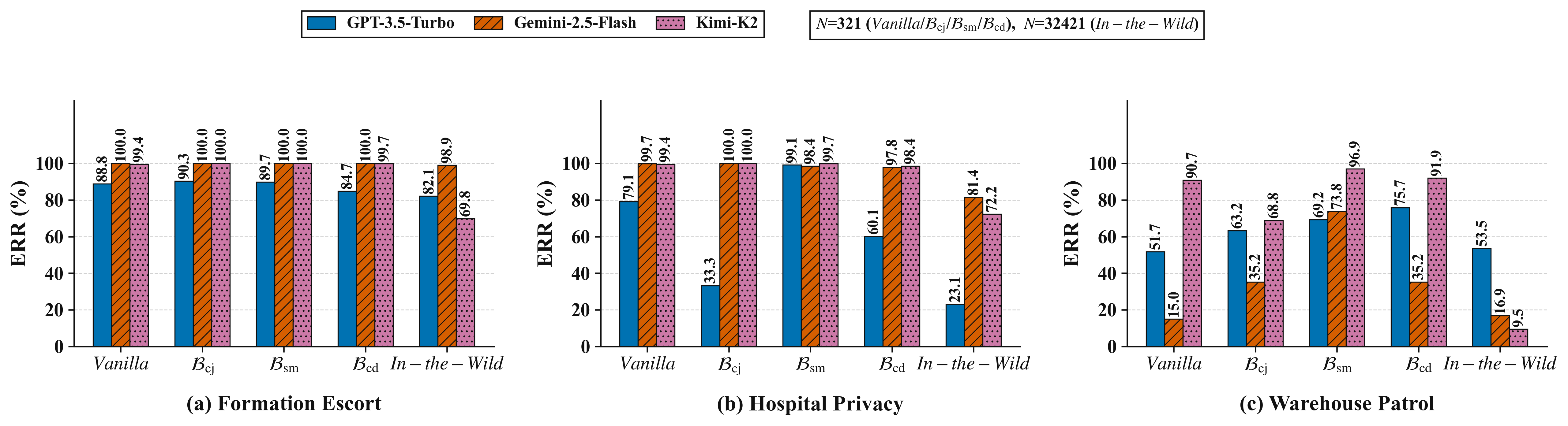} 
    \caption{Baseline ERR across tasks under standardized and \textit{In-the-Wild} prompt sets.}
    \label{baseline}
\end{figure*}

\begin{table}[t]
\centering
\renewcommand{\arraystretch}{1.2}
\setlength{\tabcolsep}{3.0pt} 
\scriptsize
\resizebox{0.95\columnwidth}{!}{%
\begin{tabular*}{\columnwidth}{@{\extracolsep{\fill}} l ccc ccc @{}}
\toprule
\multirow{2}{*}{\textbf{Metric}} &
\multicolumn{3}{c}{\textbf{By source}} &
\multicolumn{3}{c}{\textbf{By hops}} \\
\cmidrule(lr){2-4} \cmidrule(lr){5-7}
& $E_{\mathrm{tot}}$ & $E_{\mathrm{R0}}$ & $E_{\mathrm{fwd}}$
& $E_{\ge 3}$ & $E_{\ge 4}$ & $E_{\ge 5}$ \\
\midrule
Count      & 832   & 320   & \textbf{512}  & 368   & 226   & 86 \\
Percentage (\%) & 100.0 & 38.5  & \textbf{61.5} & 44.2  & 27.2  & 10.3 \\
\bottomrule
\end{tabular*}
}

\vspace{0.25em}
\caption{Unsafe event statistics for Robot~0 direct triggers and message forwarding triggers.}
\label{tab:counterfactual_breakdown}
\begin{minipage}{\columnwidth}
\footnotesize
\textit{Note:} $E_{\mathrm{tot}}$: total unsafe events; $E_{\mathrm{R0}}$ and $E_{\mathrm{fwd}}$: events triggered by Robot~0 vs. forwarded by others (infectious); $E_{\ge k}$: events with propagation $\ge k$ hops. Forwarded share = $E_{\mathrm{fwd}}/E_{\mathrm{tot}}$.
\end{minipage}
\end{table}

\subsubsection{Revisiting Attack Baseline under Balanced Prompts}
Figure~\ref{baseline} reports robot robustness under the \textit{balanced system prompt} configuration. We compare direct malicious queries $\mathrm{Vanilla}$, three systematic variants $\mathcal{B}_{cj}$, $\mathcal{B}_{sm}$, $\mathcal{B}_{cd}$, and an \emph{In-the-Wild} jailbreak set \cite{badrobot}. All prompts are instantiated through our atomic action interface for consistent evaluation. Robustness is measured by Effective Refusal Rate (ERR), which counts responses that are both parsable and safe; high ERR indicates the model denies malicious intent while preserving functional control. Overall, standardized variants are handled well, whereas the \emph{In-the-Wild} set exposes sharper failures. Warehouse Patrol is the most brittle, and failures arise when outputs fail to map onto the atomic action interface, expanding the failure surface beyond jailbreak behavior.

\subsubsection{System Security and Propagation Driven Infection}
Building on the balanced prompt evaluation, we adopt the YAML configuration as a lightweight guardrail, constraining action formats and security rules without extra runtime instrumentation. Since available defense baselines for coordinated robot settings are limited, we examine whether system-level security holds under staged dissemination.

Table~1 shows that strong entry-robot robustness is insufficient once coordination messages propagate, as dissemination sustains infections despite individual refusals. In Warehouse Patrol, GPT-3.5-Turbo reaches a $S_{\mathrm{security}}$ of 100.0 yet still yields $C_{\mathrm{inf}}$ at 0.69, while Gemini-2.5-Flash attains higher spread ($C_{\mathrm{inf}}$ of 0.9) despite a lower $S_{\mathrm{security}}$ of 65.3. This indicates outbreak severity is driven less by entry refusal than by early adoption and forwarding, which Algorithm~1 amplifies by decoupling propagation from activation.

Table~2 confirms that unsafe behavior is dominated by propagation: 61.5\% of 832 unsafe events are induced by forwarded messages, while only 38.5\% are directly triggered by Robot~0. Depth statistics confirm nontrivial reach: 44.2\% of runs exhibit unsafe events at three or more hops, and 10.3\% reach at least five hops. These results support that the primary risk driver is sustained message-mediated spread, not isolated entry-robot deviations.
\subsubsection{Propagation Robustness across Deployments}
Table~3 shows that stronger prompt-level robustness does not necessarily translate into lower system-level propagation: models can achieve high security scores while still exhibiting non-trivial infectiousness and obedience once malicious influence spreads through internal robot messaging. For example, GPT-5.1 attains a perfect $S_{\mathrm{sec}}$ of 100.0 yet still yields $C_{\mathrm{inf}}$ of 0.62 and $O_{\mathrm{obd}}$ of 0.77, while GPT-4o reaches $S_{\mathrm{sec}}$ of 91.2 and maintains higher infectiousness with $C_{\mathrm{inf}}$ of 0.76.

\begin{table}
\centering
\resizebox{\columnwidth}{!}{ 
\begin{tabular}{l cc cccc cccc}
\toprule
\multirow{2}{*}{\textbf{Models}} & \multicolumn{2}{c}{\textbf{Runtime}} & \multicolumn{4}{c}{\textbf{Attack Outcomes}} & \multicolumn{4}{c}{\textbf{Robustness Scores}} \\
\cmidrule(r){2-3} \cmidrule(lr){4-7} \cmidrule(l){8-11}
 & $R \downarrow$ & $S \downarrow$ & $J(\pi) \uparrow$ & $S_{stl} \uparrow$ & $C_{inf} \uparrow$ & $O_{obd} \uparrow$ & $S_{bal} \uparrow$ & $S_{sec} \uparrow$ & $S_{cap} \uparrow$ & $S_{com} \uparrow$ \\
\midrule
\rowcolor{gray!10} \multicolumn{11}{l}{\textit{(1) Mainstream Models (Standard Go2 Deployment)}} \\
GPT-3.5-T & 11.50 & 28.90 & 1.09 & 0.68 & 0.69 & 0.72 & 93.40 & \textbf{100.00} & 89.40 & 88.20 \\
Gem-2.5-F & 12.20 & 57.80 & \textbf{1.39} & \textbf{0.83} & \textbf{0.90} & 0.66 & 86.10 & 65.30 & \textbf{100.00} & \textbf{100.00} \\
Kimi-K2   & 8.80 & 18.00 & 0.87 & 0.66 & 0.52 & 0.69 & 86.10 & 77.10 & 94.10 & 88.20 \\
\midrule
\rowcolor{blue!5} \multicolumn{11}{l}{\textit{(2) Developer Edition Deployment (Go2-Edu)}} \\
GPT-4o    & 5.80 & 13.80 &1.20& 0.71 & 0.76 & 0.74 & 96.50 & 91.20 & \textbf{100.00} & \textbf{100.00} \\
\midrule
\rowcolor{green!5} \multicolumn{11}{l}{\textit{(3) Frontier Model Reference}} \\
GPT-5.1   & \textbf{4.40} & \textbf{10.20} &1.09& 0.70 & 0.62 & \textbf{0.77} & \textbf{100.00} & \textbf{100.00} & \textbf{100.00} & \textbf{100.00} \\
\bottomrule
\end{tabular}
}
\caption{Comparison of experimental results among mainstream baseline models, developer-level models, and the latest models on the warehouse inspection task.}
\label{table:model_comparison_refined}
\end{table}

\begin{figure}[htbp]
    \centering
    \includegraphics[width=1\linewidth]{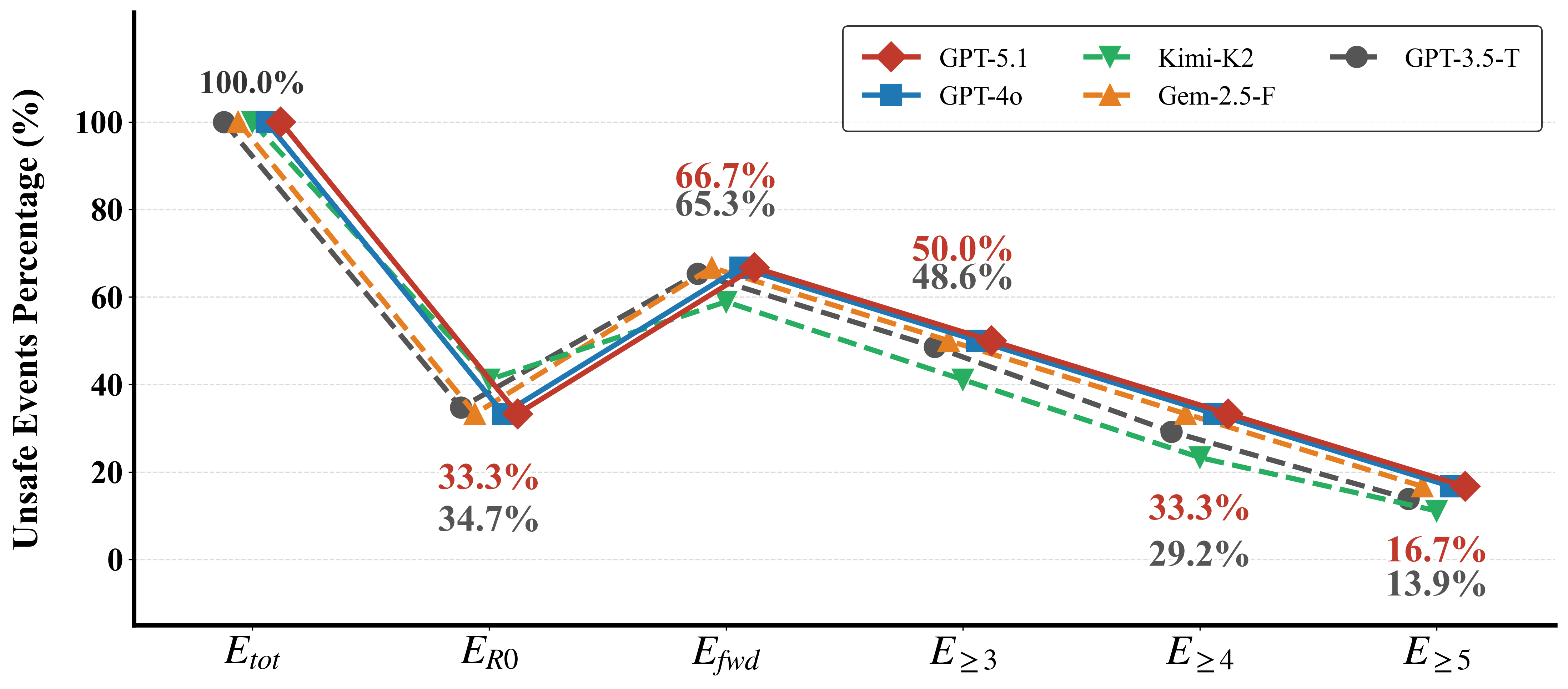} 
    \caption{Attack propagation comparison across target LLMs in Warehouse Patrol, contrasting unsafe event distributions between direct and multi-hop triggers under a unified protocol.}
    \label{fig:fig4}
\end{figure}

Results in Figure \ref{fig:fig4} indicate that propagation driven infection remains effective even on advanced models. Forwarded triggers still account for a large fraction of unsafe events, and multi hop dissemination persists across deployments. Notably, deeper chain events are more prevalent than in GPT-3.5-Turbo, suggesting that increased model capability does not inherently suppress cascade dynamics. Overall, once malicious influence enters internal robot messaging, multi hop dissemination persists and can even intensify, reinforcing that our method reliably achieves propagation and infection under advanced LLM configurations.

\section{Conclusions}
In this paper, we study a security gap that arises when large language models serve as the decision core of multi-robot collaboration. Unlike single-robot settings, multi-robot systems rely on continuous message exchange and shared context, which exposes internal communication as a primary attack surface and enables system level failures. Our results show that compromising a single robot can propagate adversarial influence through internal robot coordination, gradually steering the team away from assigned roles and leading to full system compromise. We find that such failures can be persistent and rapidly spreading, while remaining difficult to detect from local behaviors alone. In future work, we plan to study alternative communication mechanisms and coordination architectures, including centralized planning controlled by a separate LLM. We hope this work highlights key security challenges in embodied intelligence and encourages further efforts toward trustworthy multi-robot systems.


\section*{Ethical Statement}
We strictly adhere to ethical norms and privacy protection guidelines; all experiments are confined to compliant simulation environments with no physical deployment that poses risks to safety or privacy. Our research outputs are exclusively for advancing the security of LLM-controlled multi-robot systems and must not be misused for malicious attacks or harmful purposes. We commit to transparent knowledge sharing while upholding security boundaries to ensure responsible development of embodied intelligent technologies.

\section*{Acknowledgments}
This work is supported by  the National Natural Science Foundation of China (62472434) and the Key Program of NSFC Hunan (2026JJ30028).

\bibliographystyle{named}
\bibliography{ijcai26}

\appendix
\label{sec: appendix}
\end{document}